\documentclass{Interspeech}

\usepackage{multicol}
\usepackage{multirow}
\usepackage{makecell}
\usepackage{microtype}
\usepackage{cleveref}

\usepackage{tikz}
\usetikzlibrary{arrows,positioning} 
\usetikzlibrary{shapes,decorations}
\usetikzlibrary{calc}
\usetikzlibrary{fit}
\usetikzlibrary{decorations.pathreplacing}
\usetikzlibrary{backgrounds}
\usetikzlibrary{math}

\definecolor{blind_red}{HTML}{D7191C}
\definecolor{blind_orange}{HTML}{FDAE61}
\definecolor{blind_yellow}{HTML}{FFFFBF}
\definecolor{blind_blue}{HTML}{ABD9E9}
\definecolor{blind_blue2}{HTML}{2C7BB6}

\interspeechcameraready

\title{Running Conventional Automatic Speech Recognition on Memristor Hardware: A Simulated Approach}

\author[affiliation={1,3}]{Nick}{Rossenbach}
\author[affiliation={1,3}]{Benedikt}{Hilmes}
\author[affiliation={2}]{Leon}{Brackmann}
\author[affiliation={3}]{Moritz}{Gunz}
\author[affiliation={1,3}]{Ralf}{Schlüter}

\affiliation{Machine Learning and Human Language Technology}{RWTH Aachen University}{Germany}
\affiliation{Institute for Electronic Materials II}{RWTH Aachen University}{Germany}
\affiliation{}{AppTek GmbH}{Aachen, Germany}
\email{\{rossenbach,hilmes,schlueter\}@ml.rwth-aachen.de, l.brackmann@iwe.rwth-aachen.de, mgunz@apptek.com}
\keywords{speech recognition, neuromorphic hardware, memristor simulation}

\usepackage{comment}

\begin{document}

\maketitle

\begin{abstract}
Memristor-based hardware offers new possibilities for energy-efficient machine learning (ML) by providing analog in-memory matrix multiplication. Current hardware prototypes cannot fit large neural networks, and related literature covers only small ML models for tasks like MNIST or single word recognition. Simulation can be used to explore how hardware properties affect larger models, but existing software assumes simplified hardware. We propose a PyTorch-based library based on ``Synaptogen'' to simulate neural network execution with accurately captured memristor hardware properties.
For the first time, we show how an ML system with millions of parameters would behave on memristor hardware, using a Conformer trained on the speech recognition task TED-LIUMv2 as example. With adjusted quantization-aware training, we limit the relative degradation in word error rate to 25\% when using a 3-bit weight precision to execute linear operations via simulated analog computation.
\end{abstract}

\section{Introduction}
Artifical neural networks (ANN) play an important role in natural language processing (NLP) tasks such as automatic speech recognition (ASR).
The majority of current ANN architectures used in NLP such as LSTM \cite{hochreiter1997lstm} or Transformer \cite{vaswani2017attention} and related derivatives make use of tensor operations such as vector-matrix-multiplication (VMM).
VMM-based neural networks are typically executed via graphic processing units (GPU) containing many parallel processors for efficient computation of large VMM operations.
By introducing special processor units such as Tensor Cores in Nvidia GPUs or dedicated accelerators such as Google TPUs \cite {DBLP:conf/isca/JouppiK0MNNPSST23}, the efficiency further increased for many operations \cite{10.1145/3629526.3653835}.
Still, the underlying chip technology is based on complementary metal-oxide semiconductors (CMOS).

Memristor technologies offer the possibility to perform mathematical operations on analog level using a programmable but non-volatile resistance state \cite{Strukov2008}.
This is especially suited for VMMs in ANNs, as the trainable weights of the network usually stay fixed during execution.
The computation is performed in constant time by applying the input vector as voltages to a memristor crossbar that physically stores the weights as resistance, and reading the resulting currents.
This requires a fraction of the energy consumption compared to CMOS-based circuits \cite{Aguirre2024}.
A downside is the non-deterministic physical deviation of the resistance encoding.
While there is prior work discussing the scope and impact of these uncertainties, current prototype hardware often only allows storing multiple thousands of parameters \cite{Wan2022}.
Thus, implications of those uncertainties for machine learning applications were only investigated on very small tasks and with limited numbers of parameters.

In this work, we want to utilize the availability of simulation software that can mimic the stochastic behavior of an actual device.
We make use of Synaptogen \cite{Synaptogen} to develop a compact framework to mimic the execution of PyTorch-based neural networks for ASR on real memristor hardware.
While there are existing frameworks for memristor simulation \cite{Lammie2022,Rasch2021}, these make use of simplified modeling assumptions.
A detailed discussion of the modeling differences and existing frameworks will follow in \Cref{sec:Synaptogen}. Current literature covering the effects of memristor properties on ML tasks \cite{doi:10.1126/science.adf5538, souto2024neuromorphiccircuitsimulationmemristors,9108292} targets only small tasks such as MNIST \cite{lecun-mnisthandwrittendigit-2010} or CIFAR \cite{Krizhevsky09learningmultiple} for imaging and Google Speech Commands \cite{speechcommandsv2} for speech.
In addition, current literature such as \cite{doi:10.1126/science.adf5538,souto2024neuromorphiccircuitsimulationmemristors,9108292} often omits important information about how the models are trained and mapped, and focus more on the aspects of the hardware itself.
It should be noted that for our work, we assume that the memristive hardware would exist as general purpose VMM accelerator component.

The contributions of this work are as follows: We give an overview of the available memristor technology and briefly explain the consequences of the hardware properties for input-to-weight multiplications.
We present a mapping of the PyTorch nn.Linear operations to the device-based simulation of Synaptogen to simulate execution with realistic device characteristics.
In order to have a model that can be mapped, we explain our quantization-aware training scheme targeting very low-precision network weights with resolutions of down to 3 bits.
Finally, we present the results for a Conformer-based ASR system trained on the TED-LIUMv2 \cite{Rousseau2014EnhancingTT} task and show the degradations caused by the low-precision computation and the memristor hardware characteristics.
To our knowledge, this work is the first approach to discuss the influence of memristor hardware on a large scale NLP task.
The PyTorch extension for Synaptogen, the model training code and other software \cite{DBLP:conf/emnlp/PeterBN18, DBLP:conf/icassp/DoetschZVKSN17} are publicly accessible.\footnote{\scriptsize\url{https://github.com/rwth-i6/returnn-experiments/tree/master/2025-memristor-asr}}

\section{Simulated Memristor Hardware}
A memristor is a passive device whose resistance changes in response to the application of a sufficiently large electric voltage \cite{Strukov2008}.
After the voltage is removed,  the resistance of the device is conserved until a sufficiently large voltage is applied again.
By varying the voltage amplitude, the resistance of the device can be adjusted incrementally, allowing the remanent programming of multiple resistance levels in one device \cite{Yang2013}.
However, memristors exhibit an intrinsic stochastic behavior. 
Particularly the programming process remains challenging, as the resistance levels often demonstrate significant variations~\cite{Wouters2016}.
In their origin, these variabilities are comparable to the phenomena of mismatch and process variations in existing CMOS technologies and call for precise characterization.

Memristor hardware has yet to see widespread commercial availability. Semiconductor manufacturers Infineon and TSMC just recently entered the market with integrated processors featuring memristors for automotive use \cite{Infineon2022ReRAM}.

\subsection{Memristors for Vector-Matrix-Multiplication}
\begin{figure}[t]
  \centering
  \includegraphics[trim={0.75cm 0.65cm 0.75cm 0.65cm},clip,width=0.55\linewidth]{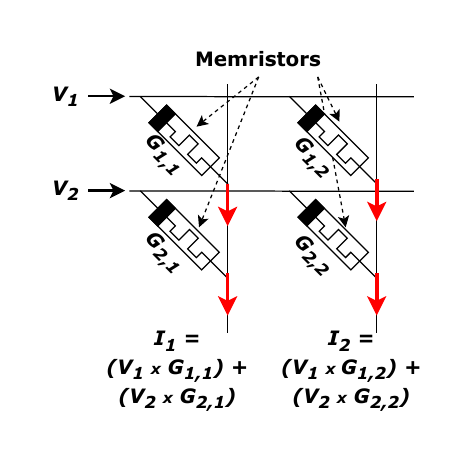}
  \caption{Schematic representation of a Vector-matrix-multiplication in a $2 \times 2$ memristor crossbar. A weight matrix is programmed as conductance level of the memristors, $G_{i,j}~(i,j\in[1,2])$. An input vector ($V_1,V_2$) is applied at the horizontal lines. The resulting current per column, $I_{1,2}$ yields the sum of the products between the inputs and the conductance. }
  \label{fig:crossbar}
  \vspace{-2.0em}
\end{figure}

Memristor crossbars are becoming a focus of research due to their ability to perform vector-matrix multiplication (VMM) directly within memory, eliminating the need for data transfer between storage and processing units \cite{Wan2022}.
Consequently, memristors are proposed as analog hardware for neural network computation, potentially augmenting or even replacing current digital processors \cite{Huang2024}.
A memristor crossbar consists of two perpendicular sets of voltage lines and a memristor device placed at each line intersection, as depicted in \Cref{fig:crossbar}.
In a memristor-based VMM, the analog values of a weight matrix, $M$, are mapped to the multiple resistance states of the memristors inside a crossbar. To better relate the resistance state to the weights of the neural network, we describe a memristor state using the conductance $G$, which is the inverse of the resistance. 
The input vector $V$ is applied as voltages along the horizontal crossbar lines.
According to Ohm's Law and Kirchhoff's current law, the current, $I$, flowing through each vertical line is the sum of the individual products of input and conductance \cite{Aguirre2024}.
Physical implementations may also contain transistors to help programming \cite{Huang2024}.
Yet, the memristor variations can induce computational inaccuracies, which impact VMM performance.
Consequently, these variability effects have to be carefully investigated to ensure reliable operation.

\subsection{Synaptogen}
\label{sec:Synaptogen}
Full realizations of memristor-based machine learning applications remain challenging due to strict hardware boundaries, e.g. the number of crossbars within a chip.
Therefore, simulation tools tailored for memristor artificial neural networks have been developed.
Prominent examples are the open-source MemTorch framework \cite{Lammie2022} and the AIHW KIT by IBM \cite{Rasch2021}.
These tools utilize the PyTorch framework to simulate ANN applications and layers based on memristor crossbars on a high level of abstraction.
However, these simulators lack accuracy in modeling the individual memristor behavior, especially the accurate definition of variability that significantly increases computational complexity. 
Although MemTorch offers several device models, including a data-driven model, the memristor parameters are idealized and the variations are based on arbitrary conductance distributions.
The variability-aware memristor model Synaptogen has been proposed to address these limitations for the accurate simulation of memristor hardware \cite{Synaptogen}.
The simulation model captures memristor variations based on vector autoregression, which was trained on measured electrical characteristics of over 3 million programming cycles.
The captured devices in Synaptogen are state-of-the-art foundry quality oxide-based memristors and commercially available based on a \SI{130}{nm} process from STMicroelectronics.
Synaptogen incorporates device and programming variations as well as physical noise sources such as thermal noise (Johnson-Nyquist) and bit quantization.
However, Synaptogen is not a high-level simulator, such as MemTorch and the AIHW KIT.
Instead, it is a mathematical device model to provide realistic hardware behavior. In order to utilize Synaptogen for large-scale ANN designs, we implemented a PyTorch extension to execute it on GPUs.

\subsection{Neural Network Operations with Synaptogen}


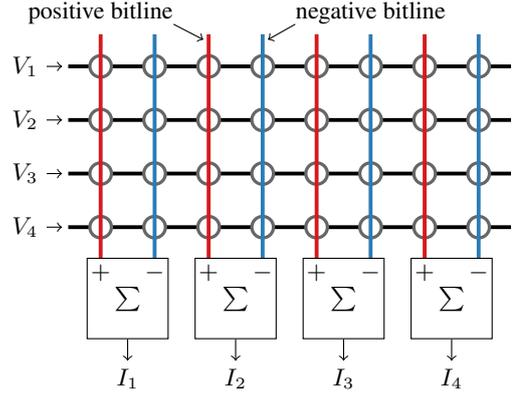
\begin{figure}
\vspace{-0.5em}
	\begin{center}
	\resizebox{2.7in}{!}{
	\begin{tikzpicture}[
	    auto,
        gridline/.style={line width=0.5mm},
        voltageline/.style={gridline, black},
        currentline/.style={gridline, blind_blue2},
        currentlineneg/.style={gridline, blind_red},
        sumbox/.style={anchor=north, rectangle, draw},
	]
    \def \scale {1.4}
    \foreach \i in {1,...,4}{
        \draw [voltageline] (\scale*2mm,\i*\scale*5mm) -- (\scale*43mm,\i*\scale*5mm);
    }
    
    \foreach \i in {1,...,8}{
        \foreach \j in {1,...,4}{
            \filldraw[color=black!60, fill=white!5, very thick](\i*\scale*5mm,\j*\scale*5mm) circle (\scale*1mm);
        }
    }
    
    \foreach \i in {1,...,4}{
        \draw [currentline] (\i*2*\scale*5mm,3mm) -- (\i*2*\scale*5mm,\scale*23mm);
        \tikzmath{\x1 = ((\i*2)-1);}
        \draw [currentlineneg] (\x1*\scale*5mm,3mm) -- (\x1*\scale*5mm,\scale*23mm);
    }

    \foreach \i in {1,...,4}{
        \tikzmath{\x1 = (5-\i);}
        \node [anchor=east] (v\i) at (0mm, \x1*\scale*5mm) {$V_\i$};
        \draw [->] (v\i) -- (2mm, \x1*\scale*5mm);
    }

    \foreach \i in {1,...,4}{
        \tikzmath{
            \sumpos = (\i*2)-0.5);
            \pluspos = (\i*2)-1);
            \minuspos = (\i*2);
        }
        \node [sumbox, minimum size=\scale*7.5mm] (sum\i) at (\sumpos*\scale*5mm, 3mm) {$\sum$};
        \node [below=\scale*2mm of sum\i] (label\i) {$I_\i$};
        \draw [->] (sum\i) -- (label\i);
        \node [] at (\pluspos*\scale*5mm,1mm) {$+$};
        \node [] at (\minuspos*\scale*5mm,1mm) {$-$};
    }

    \node [inner sep=0] (posbit) at (\scale*5mm,\scale*25mm) {positive bitline};
    \path let \p1 = (posbit.south east), \p2 = (\scale*14.5mm,\scale*22.5mm) in [->,draw] (\x1,\y1) -- (\x2, \y2);

    \node [inner sep=0] (negbit) at (\scale*30mm,\scale*25mm) {negative bitline};
    \path let \p1 = (negbit.south west), \p2 = (\scale*20.5mm,\scale*22.5mm) in [->,draw] (\x1,\y1) -- (\x2, \y2);
    
    \end{tikzpicture}
    }
    \end{center}
    \vspace{-2.0em}
    \caption{Crossbar arrangement for pairwise memristors. Inspired by \cite{Aguirre2024}.}
    \label{fig:pairwise}
    \vspace{-1.5em}
\end{figure}

Memristors can only cover the range between a specific low conductance and high conductance. To represent a weight of zero an infinite resistance would be needed so that no current flows for any applied voltage. As solution, we follow \cite{Aguirre2024} and use the pairwise difference of memristors to model a single weight as depicted in \Cref{fig:pairwise}. With the subtraction of two cells in low-conductance state, a weight of zero can be achieved. Furthermore, encoding negative weights becomes possible. Thus, we model an $N \times M$ matrix with $N$ inputs and $M$ outputs and internal states -1, 0, and 1 using an $N \times 2M$ memristor crossbar. While it is possible to also use intermediate conductance levels, those are much more imprecise to configure, as shown later in this section. The physical input range without switching the cell is $-0.6V$ to $0.6V$. In order to be able to interpret the computation done by the memristor crossbar, we need to map the resulting output currents back to the desired value space. We introduce a correction factor $c$ with the unit $\frac{1}{A}$ so that for the difference between the output currents of a cell on the positive bitline $I^+$ and negative bitline $I^-$ the following holds for any $x$ in [-1, 1]:
\begin{align}
    \left(I^+_{high}(x \cdot 0.6V) - I^-_{low}(x \cdot 0.6V)\right) \cdot c &\approx x \\
    \left(I^+_{low}(x \cdot 0.6V) - I^-_{low}(x \cdot 0.6V)\right) \cdot c &\approx 0 \\
    \left(I^+_{low}(x \cdot 0.6V) - I^-_{high}(x \cdot 0.6V)\right) \cdot c &\approx -x
\end{align}
We write $I_{high}$ for a cell in high conductance state, and $I_{low}$ for a low conductance state. For the paired memristor model in Synaptogen, the optimal $c$ w.r.t. a quadratic error was $8020\frac{1}{A}$.
Given that the memristors can not be precisely set to the desired conductance states, we obtain large deviations in the computations. Table \ref{tab:computation} shows the results for the different basic operations.
The last row shows the increased variation when trying to set the cell on the positive bitline to a state in between the low and high conductance, e.g. for achieving a weight of 0.5.
Thus, for this work we stick to binary programming of the states.
In order to model weights with higher precision, we can stack multiple crossbars, where each crossbar represents a bit level. That means for a 4-bit resolution, a first crossbar would model the weight states -4, 0, 4, the second -2, 0, 2 and the last -1, 0, 1. Thus, with a stack of 3 crossbars we can model 15 different weight levels from -7 to 7. In the physical hardware the input and output levels can not be freely chosen. In this work, we assume a resolution of 8 bit for the digital-to-analog converter (DAC) providing the input voltage and the analog-to-digital converter (ADC) reading the output currents.
As memristors only support a specific physical value range, any inputs and parameters have to be scaled with their respective quantization scales. Further explanation will be given in \Cref{sec:quantization}.
In order to fit larger matrices into the crossbars, we tile the weights across multiple crossbars of e.g. $128 \times 128$ memristor pairs, as this is a size that can be assumed to be the realistic maximum for current hardware \cite{Chen2024}. All DAC, ADC and tiling parameters are freely configurable in our framework. The memristor cell conductance state functions and additional noises are not configurable, but strictly based on the Synaptogen simulation parameters of real hardware.


\section{Automatic Speech Recognition}

\begin{table}
\caption{Computation results of 10000 single simulated cells set to perform the given multiplication in a [0,1] normalized space for input and weight. The first three rows show the non-linearity w.r.t input voltage on a single fixed device. Row \#4 shows the noise with zero weight due to the paired subtraction of two crossbar columns necessary. Row \#5 shows the increased uncertainty when trying to set "half" conductance states.}
    \label{tab:computation}
    \centering
    \setlength\tabcolsep{5px}
    \begin{tabular}{|c|l|l|l|l|l|l|}
    \hline
         operation & average & std-dev & min & max\\
         \hline
         1.0 x 1 & $1.016$ & $0.063$ & $0.707$ & $1.27$\\
         0.1 x 1 & $0.989e^{-1}$ & $0.062e^{-1}$ & $0.688e^{-1}$ & $1.24e^{-1}$\\
         0.01 x 1 & $0.985e^{-2}$ & $0.069e^{-2}$ & $0.721e^{-2}$ & $1.20e^{-2}$\\
         1.0 x 0 & $8.52e^{-4}$ & $4.75e^{-2}$ & $-0.2751$ & $0.2455$\\
         1.0 x 0.5 & $0.476$ & $0.094$ & 0.109 & 1.06\\
         \hline
    \end{tabular}
    \setlength\tabcolsep{6px}
    \vspace{-1.5em}
\end{table}

Our ASR system consists of a Conformer encoder \cite{conformer} with a Connectionist-Temporal-Classification (CTC) \cite{Graves06connectionisttemporal} output loss layer.
We chose the Conformer as encoder architecture as it is widely used in current research literature \cite{10301513}.
While most layers in a Conformer block make use of VMMs with static matrices where modeling with memristor hardware is possible, dynamic operations such as self-attention, layer-norm or gating do not have a static weight component, and are excluded from being executed on the memristor devices.
For feature extraction we use log-mel features with a shift of 10ms, followed by a stack of convolutional layers to perform down-sampling of factor 4 on the frame level.
We exclude the feature extraction and the down-sampling network from the simulated memristor execution.
Our target labels are ARPA-phonemes, and we use the official lexica provided with the dataset.
For decoding, we use the lexical prefix-tree search implemented in the open-source decoder Flashlight \cite{pmlr-v162-kahn22a}, which is accessible via a Python interface in TorchAudio.
We include 4-gram count-based language model using the KenLM \cite{heafield-2011-kenlm} interface in the recognition process.

\section{Quantization-Aware Training}
\label{sec:quantization}
Previous research utilizes already trained models, usualy using post-training quantization (PTQ) to quantize the models for further usage, as also done in MemTorch.
While PTQ works well for 8-bit precision formats, the performance degrades with decreasing bit depth \cite{gholami2021surveyquantizationmethodsefficient}.
This is because lower bit precision introduces more noise to the computations, compared to what the model sees during training. 
Nevertheless, these lower bit precisions are required for working with memristors, as they only allow for a low (or binary) precision, and stacking largely increases the number of necessary crossbars within a chip.
Quantization-aware training (QAT) describes the process of adapting the model to this quantization noise already during training \cite{gholami2021surveyquantizationmethodsefficient}.
This way, the model can learn to deal with possible inaccuracies that will occur during (lower-bit) inference.
For this, an observer is placed at the position of weights and activations during training , capturing the possible values during the current forward step. We fake-quantize each tensor into the desired precision, meaning while computations still happen in FP32, only a limited number of values are allowed.
One important property for memristor computation is the need for a symmetric quantization range with a fixed zero-point at 0.
This can be challenging for PTQ, as there is no control over the value range and the distribution the model learns during training. In contrast, QAT allows to directly induce this property during training, nudging the model to learn a symmetric distribution within the given bit precision.

\section{Simulated Execution of Conformer Blocks}
\label{sec:simulation}
During QAT, we have observers that track the statistics of each weight and input.
Assuming that for a specific linear layer we would have tracked the largest absolute input value to be 4 and the largest absolute weight value to be 0.1, we would have the input scaling factor 0.25 and a weight factor of 70 for a 3-crossbar stack with level representations from -7 to 7.
For each memristor in the simulated cell pairs representing each weight, we would apply a large positive or negative voltage to set the desired high or low conductance states.
During execution, we quantize the normalized input values with the DAC settings and apply the resulting input voltage via the cell simulation function.
After reading the results from the simulated crossbars, we apply the previously calculated correction factor $c$, apply the ADC quantization, bit-shift the results with respect to their weight bit level, and then divide both the input factor and weight factor to retrieve the desired output.
This process is performed for all parameterized linear transformations in the conformer encoder with the sole exception of the depth-wise convolution. This means that only less than 1\% of the trainable weights of the conformer blocks remain outside of the memristor crossbars.
Parameter-free operations such as activation functions, gating, the dot product of the self attention, and the softmax operations are computed as usual.
The DAC precision which corresponds to the activation quantization during QAT, as well as the ADC precision, is fixed to 8 bits.
As there is a high uncertainty of the resulting conductance when setting the weights in the memristor simulation, we report the recognition results over 10 different runs.
Between each run, we apply randomly drawn voltages multiple times to the memristor to reset the conductance states in order to mitigate correlation effects across programming cycles.

\section{Experiments}
\subsection{Experimental Setup}
We conduct our experiments on TED-LIUMv2 (TEDv2) \cite{Rousseau2014EnhancingTT}, which consists of 207 hours of TED talks. We use the dev-set of TED-LIUMv2 for evaluation. For recognition, we use a 4-gram LM trained on the text data provided with the corpus.

The model consists of 12 Conformer layers with a model dimension of 384 and a feed-forward dimension of 1536. The total number of parameters is 42M. We train the model for 50 epochs with RAdam \cite{DBLP:conf/iclr/LiuJHCLG020}, using a linearly increasing and decreasing learning rate schedule with a maximum learning rate of $5e^{-4}$. When performing QAT, we use the value range observers and the fake quantization operations right from the beginning of the training, so no non-QAT pre-training is performed. The baseline and QAT trainings are performed with identical hyperparameter settings, except for the added quantization operations.

Given that recognition with the simulation software is much slower than regular recognition, we have a real-time-factor of around 1 to 1.5 on an Nvidia L40S.

\subsection{Effect of Quantization}


\begin{table}
\caption{Post-training quantization vs. quantization-aware training on TEDv2 dev. Activations are quantized with 8-bit precision.}
\label{tab:versus}
\vspace{-0.5cm}
\center{
\begin{tabular}{|c||c|c|}
\hline
\multirow{1}{*}{\makecell{Weight\\Resolution (Bits)}} & \multicolumn{2}{c|}{WER [\%]$\downarrow$} \\
\cline{2-3}
& PTQ & QAT\\ 
\hline
\hline
8 & \phantom{0}7.2 &  \phantom{0}7.3\\
6 & \phantom{0}7.9 & \phantom{0}7.7\\
5 & \phantom{0}8.9 & \phantom{0}7.4 \\
4 & 11.4    &  \phantom{0}7.8 \\
3 & 30.7    &  \phantom{0}8.3 \\
2 & n.a.    & 22.1 \\
\hline
\hline
\multicolumn{1}{|c||}{Non-Quant Baseline} & \multicolumn{2}{c|}{7.2}\\
\hline
\end{tabular}
}
\vspace{-1.0em}
\end{table}

Our first comparison is done between PTQ and QAT for low-precision weights. Table \ref{tab:versus} shows the results for weight bit levels from 8-bit down to 2-bit. For 8-bit both approaches are on par with the non-quantized baseline. PTQ quickly degrades while QAT stays within 15\% relative degradation when going to 3-bit weights. Despite 2-bit QAT training being possible, it suffers from substantial degradation. We conclude that quantizing only after the training is not a suitable approach to prepare the model for memristor deployment.

\subsection{Effect of Simulation}

For testing the memristor simulation, we take the models that are trained with QAT using 3-bit to 8-bit weight settings.
We prepare the models by applying the steps defined in Section \ref{sec:simulation}, and then run the recognition as usual.
We do not alter any other settings and perform no specific adaptation after applying the programming voltages and letting the simulation determine the cell conductance. This is in contrast to other work, where the trained networks are specifically tuned towards the resulting conductance states \cite{doi:10.1126/science.adf5538}. Table \ref{tab:simulation} shows the results when taking the QAT trained models and executing recognition with the memristor model. One can see that the deviation from the baseline is substantial, but still in a reasonable range, given that no particular adjustments w.r.t. the hardware were made. There is a noticeable difference in WER for each specific programmed instance of the device, but the outliers do not exceed 5\% relative degradation compared to the mean. 
Thus, we show that it is possible to map a Conformer model trained with QAT to a realistically modeled memristor device and achieve reasonable WER.

\begin{table}[t]
\caption{Results of Conformer CTC recognition on TEDv2 dev when using the simulated memristor hardware. Results are averages over 10 differently drawn devices, including the standard-deviation as well as minimum and maximum over the runs.}
\label{tab:simulation}
\vspace{-1.0em}
\center{
    \begin{tabular}{|c|c|c|c|}
    \hline
\multirow{2}{*}{\makecell{Weight\\Resolution (Bits)}} & \multicolumn{3}{c|}{WER [\%]$\downarrow$} \\
\cline{2-4}
& avg $\pm$ std & min & max \\
\hline
\hline
8 & $8.3 \pm 0.13$ & 8.1 & 8.5\\
6 & $8.3 \pm 0.09$ & 8.2 & 8.5\\
5 & $8.3 \pm 0.16$ & 8.1 & 8.6\\
4 & $8.9 \pm 0.12$ & 8.7 & 9.2 \\
3 & $9.2 \pm 0.12$ & 9.0 & 9.4\\
\hline
\hline
Non-Quant Baseline & \multicolumn{3}{c|}{$7.2$}\\
\hline
\end{tabular}
}
\vspace{-1.5em}
\end{table}

\section{Limitations and Future Work}

This work only investigates the effect of a particular kind of memristor hardware on linear transformation layers for a single speech recognition system.
As chip co-integration has been disregarded in this work, we can not make any assumptions yet about the speed and energy benefit achieved by using memristor hardware.
In the future, we would like to focus on aspects that are relevant for the direction of research regarding the hardware capabilities. This would include research about programming the memristors with more than 2 states as well as a more detailed investigation about the needed ADC and DAC resolutions and ranges.
Additionally, we would like to extend the current implementation to further network components such as convolutions or recurrent units such as LSTMs \cite{hochreiter1997lstm}.
Through this, we can investigate the behavior of further ASR architectures such as Transducers \cite{graves2012sequence}. 
Outside a simulation, the maximum of allowed model parameters is given by the crossbar count of a specific memristor chip and the used mapping method. 
Future work should investigate the optimal trade-off between layer count, model size and necessary bit resolution either via multi-conductance states or crossbar stacking.

\section{Conclusion}

In this work, we presented how a Conformer-based ASR model performs on simulated memristor hardware that introduces a strong uncertainty in tensor operations.
We present a PyTorch extension to the Synaptogen toolkit to easily map nn.Linear layers in order to allow neural networks to be executed via a realistic memristor device simulation.
We described a detailed approach on how to simulate the execution of a Conformer ASR system on a memristor based device.
We could show that even under strong weight deviations and non-linear behavior, the system can perform reasonably well using solely QAT and no further hardware-specific tuning.
Given the ability to place a sufficient number of memristor crossbars on a single chip, the current precision of memristor crossbars is already sufficient when being used in a binary-paired setting.
While there are many open questions on how actual hardware would perform in terms of energy and speed, we can show that the non-deterministic behavior of memristors poses no restrictions on replacing fixed-weight linear operations with memristive counterparts.
Nevertheless, many issues regarding scaling the number of crossbars in a chip and the co-integration with regular computation devices have to be solved in the future.

\section{Acknowledgments}

This work was partially supported by NeuroSys, which as part of the
initiative “Clusters4Future” is funded by the Federal Ministry of
Education and Research BMBF (funding IDs 03ZU2106DA and 03ZU2106DD).
\bibliographystyle{IEEEtran}
{\footnotesize \bibliography{mybib}}

\end{document}